# Deep Neural Network and Data Augmentation Methodology for off-axis iris segmentation in wearable headsets

Viktor Varkarakis[1], Shabab Bazrafkan[2], and Peter Corcoran[3]

## Abstract

A data augmentation methodology is presented and applied to generate a large dataset of off-axis iris regions and train a low-complexity deep neural network. Although of low complexity the resulting network achieves a high level of accuracy in iris region segmentation for challenging off-axis eye-patches. Interestingly, this network is also shown to achieve high levels of performance for regular, frontal, segmentation of iris regions, comparing favorably with state-of-the-art techniques of significantly higher complexity. Due to its lower complexity this network is well suited for deployment in embedded applications such as augmented and mixed reality headsets.

## 1. Introduction

Biometric user authentication is available on consumer devices, including smartphones, using facial recognition [1]–[3] and fingerprint biometric [4]–[9]. The broad adoption of biometrics on consumer devices was originally discussed in [10] with additional discussion of the impacts in several following articles [11]–[13]. Being a near ideal biometric, the iris of the human eye is well-suited to many consumer applications, but iris recognition is traditionally implemented in a controlled environment and under constrained acquisition conditions.

Authentication requirements in consumer devices are evolving beyond today's mobile devices. New virtual reality (VR) and augmented reality (AR) headsets provide a gateway to sophisticated virtual worlds and online services [14]–[16]. In fact researchers have been working with Augmented Displays for more than 20 years [17]–[23]. The most recent mass market experiment with a wearable, augmented/mediated-reality display, that could be worn on a day-to-day basis, was Google Glass [22], [24], [25]. *Glass*, as it became known, was considered to be a game changing technology for a few years across a wide range of industry sectors [26]–[29]. But ultimately, the product was withdrawn [30].

A key challenge with AR/VR headsets is that, lacking a physical keyboard they do not provide an intuitive mean of user authentication. The weak authentication available in *Glass* [25], [31], [32] subsequently led to various attempts to refine and improve on the basic authentication of the headset [33]–[35]. Ultimately, the device authentication was simply not adequate and led, in part, to its withdrawal from the market.

This leads us to consider how the next generation of wearable AR/VR vision systems might implement a more seamless and intuitive authentication mechanism without sacrificing security and robustness. The implementation of a face recognition system is not practical, as the form-factor of an AR/VR head-set does not allow to capture a full facial image. However, with the reduction in size and cost of multi-cameras systems on mobile devices it is now practical to consider that rear-facing (i.e. user-facing) camera systems can be incorporated into such headsets. One important driver for rear-facing cameras is the use of eye-tracking to dynamically determine the wearer's point of gaze (PoG) which is important for accurate AR/VR rendering [36], [37].

---

[1] With the Department of Electronic Engineering, College of Engineering, National University of Ireland Galway, University Road, Galway, Ireland. (E-mail: v.varkarakis1@nuigalway.ie ).
[2] With imec-Vision Lab, Department of Physics, University of Antwerp, Antwerp, Belgium (E-mail: shabab.bazrafkan@uantwerpen.be )
[3] With the Department of Electronic Engineering, College of Engineering, National University of Ireland Galway, University Road, Galway, Ireland. (E-mail: peter.corcoran@nuigalway.ie ).

Iris authentication is a proven and reliable biometric trait with high distinctiveness, permanence and performance [38]. The use of iris recognition on consumer devices is explored across multiple works [39]–[42] and the importance of accurate iris segmentation, particularly in consumer imaging devices, is identified as a key challenge [43], [44]. In the iris authentication workflow, failed segmentations represent the single largest source of error [45]–[47]. In addition to its role in improving the performance of an iris-based authentication system, the accurate segmentation of iris regions, can be used successfully for eye-gaze estimation [48]. Eye-gaze as mentioned is a key element of various user-interface modalities for wearable AR/VR displays.

## 1.1. Background to the Problem

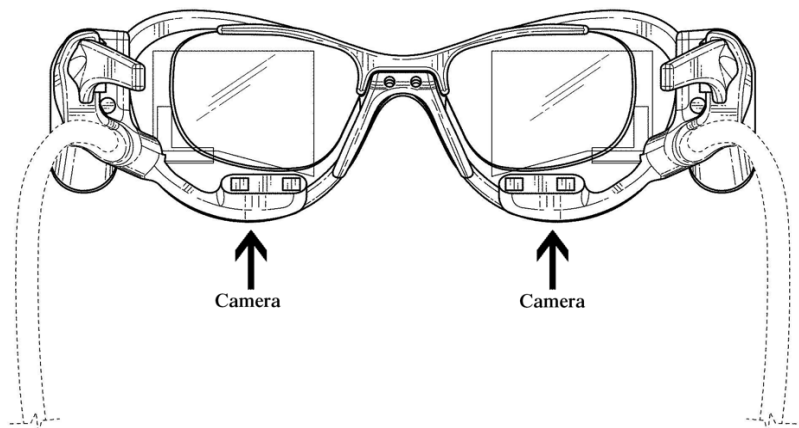

*Figure 1: Virtual Reality Glasses – Patent Number: US D795952S1 [49]*

As mentioned in the introduction, the next generation of wearable AR/VR vision systems will have to implement a more seamless and intuitive authentication mechanism that is available on today's mobile devices. US design patent, D795952S1 [49] Figure 1, shows an example of how the next generation of AR/VR headsets might incorporate a user-facing camera for iris authentication or eye-gaze tracking. Figure 2 [50][51] shows several alternative camera attachments possibilities for contemporary AR/VR devices – these are currently targeted for eye-tracking. In all these examples off-axis iris images can be readily obtained and provide a suitable biometric for user authentication.

Note that a key challenge for accurate iris recognition is to accurately segment the iris region [43], [52]. Given that camera locations for AR/VR devices must be mounted off-axis, often with an oblique perspective and close proximity to the observed eye-region the segmentation process for such off-axis iris regions becomes even more critical as errors at the segmentation stage are propagated to the feature extraction and pattern matching stages of the authentication workflow [43], [45], [53]. While there are past studies on off-axis iris and its effects on recognition rates, the problem of near-eye iris segmentation is a new problem arising from the introduction of emerging AR/VR headset technology into consumer devices.

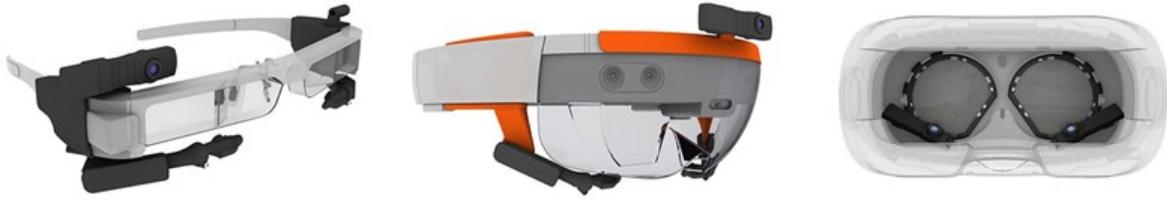

Figure 2: User-facing camera attachment system in current AR/VR systems [50]

The majority of existing iris recognition systems follow the authentication workflow as (i) image acquisition: an eye image is acquired using a camera, (ii) iris segmentation: eye/iris region is located in this image followed by isolating the region representing iris. (iii) iris normalization (iv) Feature extraction: relevant features which represent the uniqueness of the iris pattern is extracted from the iris region and (v) similarity of the two iris representation is evaluated by pattern matching techniques. The described workflow is illustrated in Figure 3, highlighting the focus of this work, which is on the second step of the iris recognition workflow, the iris segmentation. Specifically, a deep learning technique is implemented to successfully segment off-axis close proximity iris images as represented when captured from a user-facing camera on an AR/VR device.

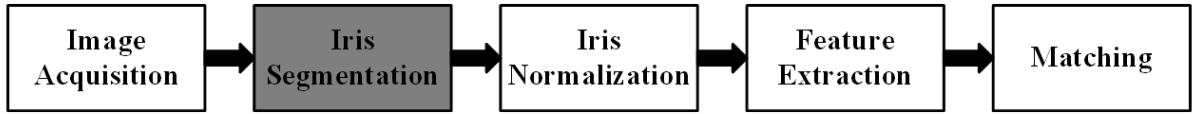

Figure 3: Iris authentication workflow. In practical implementation the bulk of authentications errors are due to incorrect segmentations [45][46][47].

## 1.2. Related literature

As background a quick overview of iris segmentation works in the literature is outlined below.

### 1.2.1. Frontal Iris Segmentation

The number of methods in the literature regarding iris segmentation shows that this topic has been thoroughly studied but remains an active area of research. When referring to iris segmentation algorithms, a good starting point are two highly cited works in the literature: Daugman [54] and Wildes [55]. In the iris matching algorithms developed in these research papers iris segmentation is achieved by fitting a circular contour to the iris and pupil. These two methods differ mostly in the way they define the circular boundaries on the image information. Daugman's integrodifferential operator searches the entire image pixel by pixel to find the best circular path for the iris and pupil boundaries. While Wilde, in order to fit the circular contour, combines an edge detector and Hough transform.

In continuance approaches were implemented in an attempt to speed up the process. For example, Liu [56], uses a Canny edge detector with a Hough transform to provide a fast localization of the iris edges with the assumption that the iris texture is located between two homocentric circles. Several other methods were developed based on Wilde's and Daugman's implementations such as: Huang's [57], Khan's [58], He's and Shi's [59], Lili's and Mei's [60].

As noted, the aforementioned methods assume the circularity of the iris outer boundary and pupil boundaries. However, Daugman in his follow up work [61] shows that a non-circularity applies to the iris and pupil contour which when defined precisely it has an significant influence on recognition performance. Therefore, adopts an active contours or snake model to segment the iris. Furthermore, Shah [62] implemented a geodesic active contour to capture the iris texture and experimental result on non-ideal iris images designate the effectiveness of this method. Koh [63] similarly implemented an active contour model which was combined with the Hough transform for iris localization. In another approach, Broussard [64], used a feature saliency algorithm to identify the measurements that could define the iris boundary. The selected measurements are fed to a shallow artificial neural network in order to accurately predict the outer iris boundary. A detailed overview of the iris segmentation literature can be found in [65],[66] as well as in [67] where approaches for segmenting non-ideal iris images are reviewed.

### 1.2.2. Off-axis Iris Segmentation

A subsection of non-ideal iris images includes the off-axis iris images. Localizing the iris in this type of images has always been a challenge for researchers. In [68], Dorairaj assumes that a rough estimation of the angle rotation is available in order to deal with the off-axis iris problem. Two different objective functions are used to refine the estimate. When two images are available from the same iris class, the "ideal" and off-axis iris image, the Hamming distance between the ICA coefficients of the two images is calculated. In the case that only the off-axis image is available, Daugman's integro-differential operator is used. A projective transformation is applied to rotate the off-axis image into a frontal view image once the angle is estimated. In the next step, the image is enhanced and segmented with the integro-differential operator. In another approach, Li in [69] first fits an ellipse to the pupil boundaries. After that based on the information that has been retrieved from the ellipse fitting, rotation and scaling are applied to the image, to restore the straight position of the ellipse and the circularity of the pupil. The segmentation of iris is then operated by Daugman's like algorithms. A similar approach can be found in [70] where the use of projective and affine transformation is explored in order to bring the off-axis iris images and match them with frontal iris images. This approach comes with some serious downsides, such as the blurring of the iris outer boundaries and the fact that a prior knowledge of the angle is required for the transformation. Finally, in [71] the use of active shape models to retrieve the elliptical boundaries of the off-axis iris is investigated.

### 1.2.3. Deep Learning Approaches for Iris Segmentation

Liu, in [72] proposed two CNN approaches to segment noisy iris images acquired under unconstrained conditions. In the first approach called hierarchical convolutional neural networks (HCNNs), three patches taken from different scales of the same image are used as input. The HCNN consists of three similar blocks, a combination of convolutional and pooling layers that are merged together into a fully connected layer. In the second approach, 31 convolutional layers and 6 pooling layers are used to compose the multi-scale fully convolutional network (MFCNs). Both models are end-to-end, with no requirement for pre- or post-processing of the image. Arsalan [73], introduced a two-stage iris segmentation method. The first stage includes a pre-processing of the image and the use of a modified Hough Transform to identify the region of interest (ROI). In the second stage, a mask of $[21 \times 21]$ pixels, based on the ROI defined in the previous stage, is fed to a pre-trained VGG-face model which classifies the pixels as iris or non-iris. In a follow up work which is focused on segmenting low quality iris images, Arsalan in [74], proposed a densely connected fully convolutional network (IrisDenseNet), consisting of two main components: a densely connected encoder and a SegNet decoder. In a similar work, Bazrafkan in [43], presented a network design focused on segmenting iris of inferior quality. Four different end-to-end fully convolutional networks are merged into a single model using a method known as Semi Parallel Deep Neural Networks (SPDNN). In this way, the final model benefits from each of the four distinct network designs. Finally, since the existence of a large labelled dataset is a prerequisite in order to implement a convolutional neural network approach, Jalilian in [75] to overcome this obstacle, introduced a domain adaption method so that a CNN for iris segmentation could be trained with a limited data.

## 1.3. Contributions

The focus of this work is to improve the segmentation of off-axis iris images originating from the unconstrained conditions of a user-facing camera on wearable AR/VR device.

The model proposed is an end to end deep neural network which accepts an off-axis eye-region image and generates the corresponding binary segmentation map for the iris region as output. Performance evaluation of the proposed model shows advantages over recent iris segmentation techniques in the literature which together with its simple, yet efficient design makes it well-suited for deployment in wearable AR/VR devices.

Three noteworthy contributions are presented in this work.

1. The main contribution is a low complexity neural network design for the iris segmentation task with reduced memory requirements and computational requirements in comparison with other deep learning iris segmentation techniques.
2. A data augmentation technique that generates distorted iris images of size and quality typical of the user-facing camera employed on today's wearable AR/VR headsets. These are derived from a high quality iris dataset together with a corresponding ground truth.
3. A thorough evaluation of the proposed segmentation model is presented on several well-known public iris datasets. The presented method is compared with state-of-the-art iris segmentation techniques.

## 1.4. Foundation Methods
### 1.4.1. Network Design

The main contribution of this work is a low complexity network used for generating the segmentation map for low quality off-axis iris images.

In order to achieve high performance results, when a network is designed, large structures with high capacity are favoured. That is translated into CNNs containing millions of parameters, which to be used require large memory and high operation cost. Therefore, executing deep CNNs requires significant hardware resources which is a limited specification in many computational platforms.

The number of parameters in the proposed network is significantly lower compared to the parameters of other deep learning approaches designed for the iris segmentation task. Thus, making the proposed network faster and with reduced memory requirements, while attaining high performance results in producing the segmentation map for off-axis iris images of low quality as represented when captured by a user-facing camera on AR/VR headset and therefore well-suited for deployment in such devices.

### 1.4.2. Data Augmentation

Data augmentation is a common technique in Deep Learning exploited by researches in order to overcome the obstacle of limited labelled data. In addition, with the appropriate data augmentation techniques one can introduce variation to the training samples which results in reducing the overfitting during training but also increases the generalization of the trained network. Common techniques for the data augmentation task involve rotation, translation, flipping or adding noise.

Despite the large amount of data available today, there are still situations where either only small datasets can be found or there aren't any made publicly available. In some cases, the data contains sensitive information such as in medical applications or due to privacy / legislations reasons, data is not easily accessible. Also, with the technology rapidly growing, new problems arise frequently and in many cases it takes a while before a proper dataset is built and made publicly available. The investigated challenge in this work is a clear example of the later situation. Therefore, data augmentation is utilised to overcome the non-availability of data due to the aforementioned reasons. Applying the appropriate augmentation techniques to available datasets allows to investigate a problem for which there are no dataset assigned for. Through data augmentation one can simulate the features and characteristics of these problems without going through a data acquisition process and thus eliminating the limiting factor of the unavailability of datasets related to the problem.

In regard to our problem the main focus of the augmentation is to simulate off-axis iris images as captured by a user-facing camera on AR/VR device. As shown in Figures 1-2, a possible location of the user-facing camera utilized for iris recognition and eye-gaze is below the eye. In that case the iris samples obtained will be off-axis in the horizontal and the vertical plane. Main characteristics of the iris images taken with a head-mounted device are their elliptical shape along with the fact that are not centred, in contrast with frontal iris images where the iris is most of the time centred and a more circular shape is obtained. Therefore, the augmentation techniques in this work are focused in achieving the described representation. In addition, a secondary goal of our augmentation is to introduce effects of images when captured in real-life, where the samples are not obtained in constrained conditions and a lower quality is reported. These augmentation techniques are focused on reducing the contrast between the iris and the pupil as well as adding noise to the samples.

The rest of the paper is arranged as follows: In section 2, the datasets used are presented along with a detailed description of the augmentation techniques. In section 3, the network design and training are explained. In section 4, the results are illustrated and in section 5, the numerical evaluation of the proposed method and its comparisons with state-of-the-art segmentation techniques are presented.

Finally, it should be remarked that preliminary results from this research were first presented in [76]. This article builds on that earlier work with more detailed and extensive experimental verifications, exhaustive description of the augmentation techniques and direct comparison on off-axis and frontal iris samples with state-of-the-art iris segmentation techniques.

## 2. Datasets and augmentation methodology overview

In this work, three datasets are utilised. CASIA Thousand [77] and Bath800 [78] are used during the training and testing stages. UBIRIS v2 [79] is used for tuning and testing. Two types of augmentation methods are described below. The first type is concentrated on adding real-world condition effects to the iris images, while the second is focused on augmenting the images so that they represent off-axis iris images. The combination of these two types of augmentation methods results in iris images as captured by an user-facing camera on AR/VR device. Below, the datasets used are presented along with the production of their ground truth, and finally, the augmentation techniques are explained.

### 2.1. Datasets

CASIA-Iris-Thousand is a subset of CASIA-Iris V4 dataset. This subset contains 20000 iris images from 1000 subjects. The iris images are constrained, high quality and high contrast. Bath800 dataset is made of 31997 images taken from 800 individuals. The samples similarly to the CASIA Thousand are of high quality and high contrast. Both datasets consist of Near InfraRed (NIR) samples. Finally, UBIRIS v2 dataset includes 11102 iris images from 261 subjects, captured in visible wavelength. The samples are of low-quality as they are taken under unconstrained conditions. More detailed description of CASIA Thousand, Bath800 and UBIRIS v2 can be found in [43]. Samples from the datasets used in this work are shown in Figures 4-6.

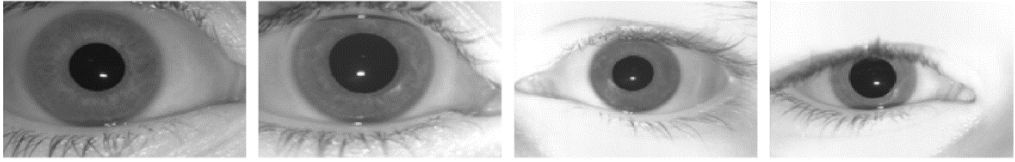

Figure 4: Eye socket samples from Bath800 dataset

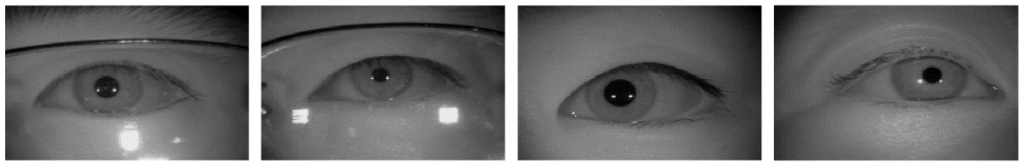

Figure 5:Eye socket samples from CASIA Thousand dataset

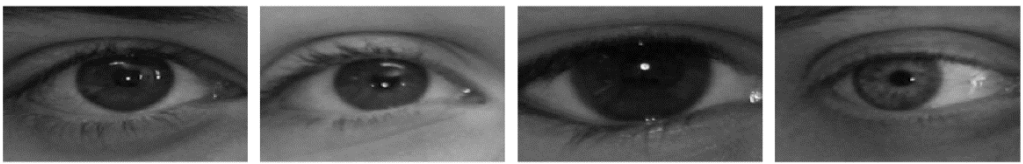

Figure 6:Eye socket samples from UBIRIS v2 dataset

## 2.2. Ground Truth

Bath800 and CASIA Thousand are not provided with the segmentation ground truth. However, these datasets as mentioned above contain images of high quality, high contrast and are captured under constrained conditions. In this work, the binary iris map for these datasets is produced using the commercial iris segmentation solution MIRLIN [80]. The obtained segmentation map is considered in this work as the ground truth. The selection of the segmentation algorithm is based on the availability as well as its performance on large-scale iris evaluations [81]. The same segmentation solution was also adopted in [43]. The low-resolution segmentations for Bath800 and CASIA Thousand are publicly available[4].

Regarding UBIRIS v2, the manual segmentation generated by WaveLab[5] [82], available in IRISSEG-EP dataset [53], is used. The manual segmentation map is not available for all the samples of the dataset. Segmentation of only 2250 images from 50 individuals is provided and therefore only these are used in this work. Segmentations examples derived from these datasets are shown in Figures 7-9.

---

[4] https://Goo.gl/JVkSyG.

[5] http://www.wavelab.at/sources/Hofbauer14b/

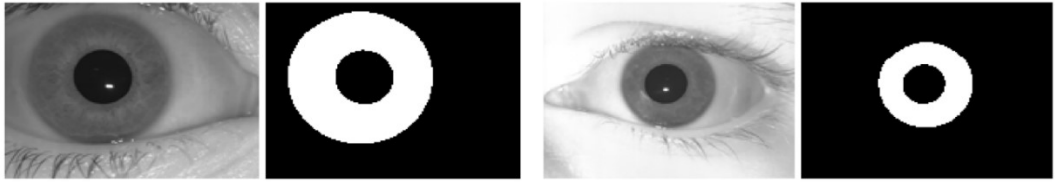

*Figure 7: Bath800 automatic segmentation results*

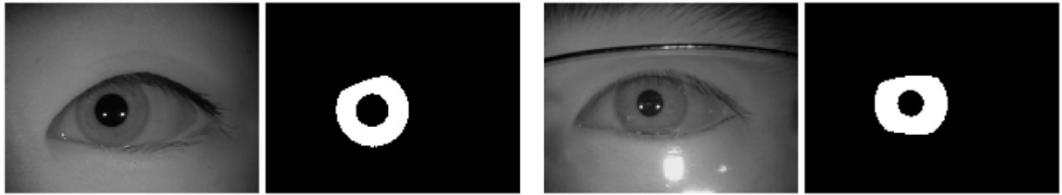

*Figure 8: CASIA Thousand automatic segmentation results*

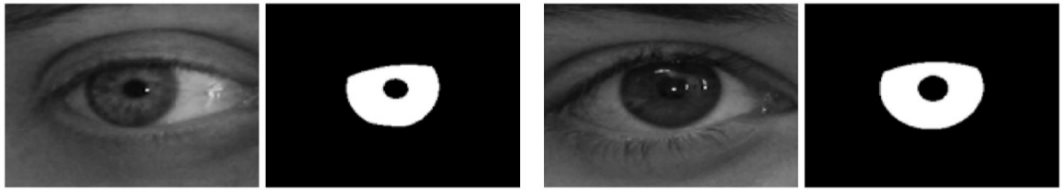

*Figure 9: UBIRIS v2 manual segmentation results*

## 2.3. Data Augmentation

In order to accurately train a deep neural network, a large number of labelled training samples are required. These samples should correctly characterize the imaging problem so that it enables the deep learning process to train an accurate model. Even if such datasets were available it would require an accurately marked ground truth – a task which poses new problems over more conventional frontal iris images. Thus, in order to obtain a large number of samples to enable the training of a DNN for AR/VR iris segmentation task, some specialized augmentations of existing datasets are required. To find the best augmentations for the iris images, precise observations have been made on iris images obtained by a user-facing camera on head-mounted displays.

The augmentation techniques are divided into two categories. The first category of augmentation techniques is focused on representing real-life scenarios where low-quality images are obtained. Based on research that has been done in [40][83], the difference between high-quality constrained iris images and wild ones is linked to contrast, blurring, and shadows. Consequently, to simulate the effects of real-world conditions in iris images the contrast is changed, motion blurring and shadows are added to the images. The augmentation techniques used to deteriorate the image quality and simulate unconstrained conditions are derived from [43]. The objective of the second category's augmentation techniques is to simulate the representation of iris images as captured by a user-facing camera on an AR/VR device. This representation includes off-axis iris images mainly of elliptical shape and not centred in the image.

The augmentation techniques are detailed in the sections 2.3.1 and 2.3.2. The workflow that is followed for the augmentation of the datasets is described in the section 2.3.3. In this work all the samples are resized to [120 × 160] using bilinear interpolation. Smaller resolution samples are preferred rather than larger ones as it accelerates the training of the deep neural network.

### 2.3.1. Data Augmentation: Simulating unconstrained conditions

The first type of augmentation techniques is applied to ensure that the samples used to train the network represent real-life scenarios. The distribution of the input data plays a vital role in what the network learns and how it will behave during the testing stage but also in unconstrained situations. As mentioned earlier, to simulate real-life captured iris images of low quality, the contrast of the samples is changed, blurring and shadows are added to the samples with the following augmentation techniques. The techniques mentioned below are derived from [43] and used with slight changes. The original code of these augmentation techniques is available.[6]

#### 2.3.1.1. Augmentation 1: Image Contrast

The iris images captured by an AR/VR device in real-world conditions compared to the high-quality, high-resolution NIR iris images acquired in constrained conditions have significant differences. The differences are with regard to the amount of contrast inside and outside the iris region as in unconstrained scenarios the samples are suffering from low contrast. Another difference noted is the intensity properties of the low-quality samples inside and outside the iris region. The region inside the iris is darker than the same region in high-quality samples. For the outside region of the iris the level of brightness cannot be categorized as it could differ from overexposed and strongly bright till very dark. To bring these properties to high-quality images, the contrast inside and outside the iris region is modified separately. This is achieved with the use of histogram mapping. The following histogram mapping equations are used to reduce the contrast of the iris images. The equation (1) is used for the region outside the iris and (2) is used for the region inside the iris.

$$y_{out} = norm(\tanh\left(3 \times \left(\frac{x}{255} - 0.5\right)\right) + \mathcal{U}(-0.2, 0.3))) \times 255 \tag{1}$$

$$y_{in} = norm(\tanh\left(3 \times \left(\frac{x}{255} - 0.45\right)\right) - \mathcal{U}(0, 0.2))) \times 255 \tag{2}$$

Where x is the input intensity in the range [0,255], $y$ is the output intensity in the same range, $\mathcal{U}(a, b)$ is the Uniform distribution between $a$ and $b$, and the norm function normalize the output between 0 and 1. As mentioned above the outside and inside region of iris suffers from low contrast, but the brightness differs. For the region outside of the iris, the histogram mapping with the equation (1) can result is bright, dark, or normally exposed low contrast outputs. For the region inside the iris, where the equation (2) is used, the contrast is reduced while the brightness of the iris region is reduced as well. Different equations are used to reduce the contrast in the inside and outside region of the iris so that variety is obtained. An example of this step is shown in Figure 10.

#### 2.3.1.2. Augmentation 2: Motion Blur

Wearing AR/VR devices, head movements are inevitable. These movements can cause motion blur. Therefore, to mimic these situations and train the model in order to be efficient in these cases, motion blurring has to be introduced to the training images. In order to include this effect, the image is passed through a motion blur filter, applying the linear camera motion by $\mathcal{U}(3,7)$ pixels in the direction $\mathcal{U}(-\pi, \pi)$, where $\mathcal{U}(a, b)$ is the Uniform distribution between $a$ and $b$. The low contrast image after applying motion blur is shown in Figure 11.

---

[6] https://github.com/C3Imaging/Deep-Learning-Techniques/blob/Iris_SegNet/DBaugmentation/DBaug.m

### 2.3.1.3. Augmentation 3: Shadowing

In unconstrained conditions, the illuminations scenarios vary. One main effect produced by different illumination directions is shadows. In order to add this effect, the iris images were multiplied with the following shadow function:

$$y = norm\ (\tanh(2 \times randSign \times (x - 0.5 + \mathcal{U}(-0.3, 0.3)))\ ) + \mathcal{U}(0, 0.1) \qquad (3)$$

where x is the dummy variable for image column number and y is the coefficient for intensity, $\mathcal{U}(a, b)$ is the Uniform distribution between $a$ and $b$, the norm function normalizes the output between 0 and 1, and the $randSign$ generates a random coefficient in the set $\{-1, 1\}$ which determines the direction of the shadow. The final image after applying shadowing is given in Figure 12.

The segmentation map for these augmented samples is the same as the original segmentation ground truth as the structure and position of the iris remains unchanged. More detailed information regarding the augmentation techniques simulating unconstrained conditions can be found in [43] from where are originated.

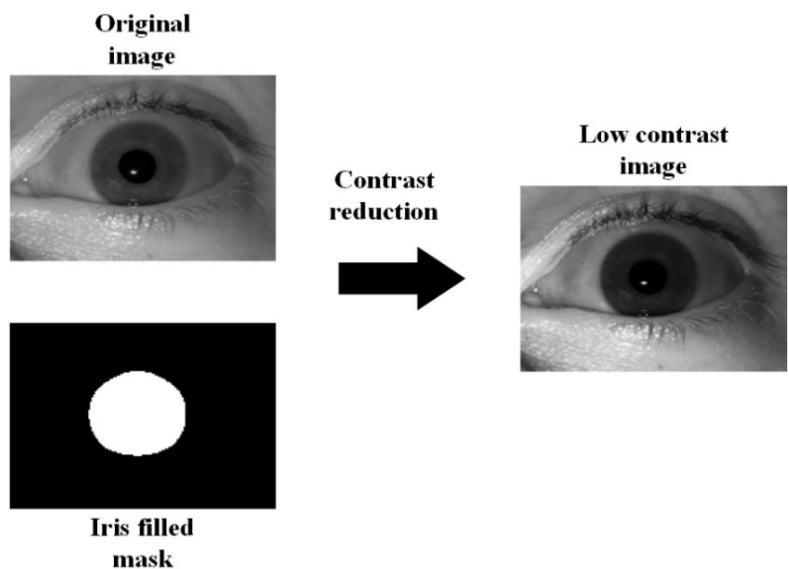

*Figure 10: For inside the iris region, the contrast is reduced, and the region is getting darker. The outside of iris is altered by decreasing the contrast.*

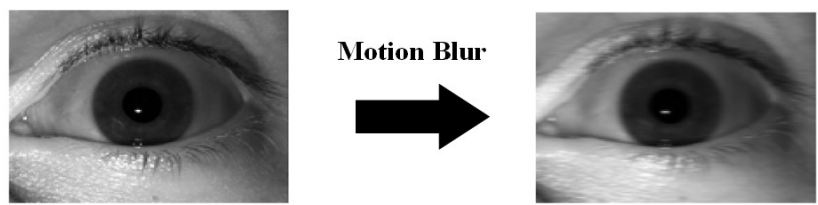

*Figure 11: Applying motion blur in a random direction to the low contrast image.*

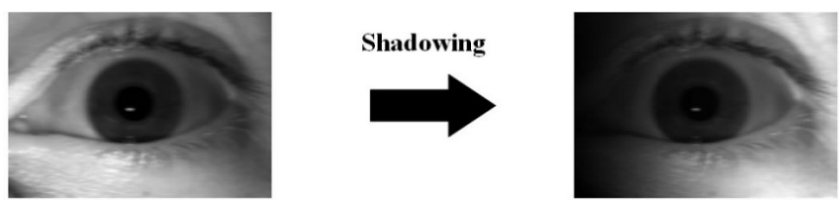

*Figure 12: Shadowing applied to low contrast blurred image.*

## 2.3.2. Off-Axis, Near-Perspective Iris Data Augmentation

The second category contains two augmentation techniques, which their goal as mentioned previously is to generate iris images as they appear when acquired by an user-facing camera on AR/VR device. As noted in the introduction and illustrated in Figures 1-2, a possible location of the camera used for obtaining an iris image that is to be used in iris recognition or eye-gaze is below the eye. Therefore, the iris images captured are off-axis in both horizontal and vertical plane. The augmentation techniques described below are specialized to produce such off-axis iris images. The code for these augmentation techniques is also available[7].

### 2.3.2.1.   Augmentation 4: Spatial stretching/contracting

The iris images when captured from an AR/VR, are characterized as distorted and with an elliptical shape. In addition, the iris is not at the centre of the image as usual. In order to generate iris images with these properties, the samples are warped by applying a spatial stretching/contracting to the iris images. The stretching is linearly applied to the images. The stretching is achieved by mapping every column/row of the image to a new position given from $y[j]$ as shown in Figure 13.

The equations below illustrate how $y[j]$ is calculated for columns/rows:

$$\lambda = \mathcal{U}(2,17) \tag{4}$$

$$k[i] = \frac{\left(\frac{1}{\lambda}\right) - \lambda}{s-1} \times t[i] + \lambda \, , i \in [1, s] \tag{5}$$

$$k[i] = \frac{\frac{(\lambda - 1)}{\lambda}}{s-1} \times t[i] + \lambda \, , i \in [1, s] \tag{6}$$

$$a = a + k[j] \, , j \in [2, s] \tag{7}$$

$$y[j] = Round(a * 5), j \in [2, s], \; y[1] = 1 \tag{8}$$

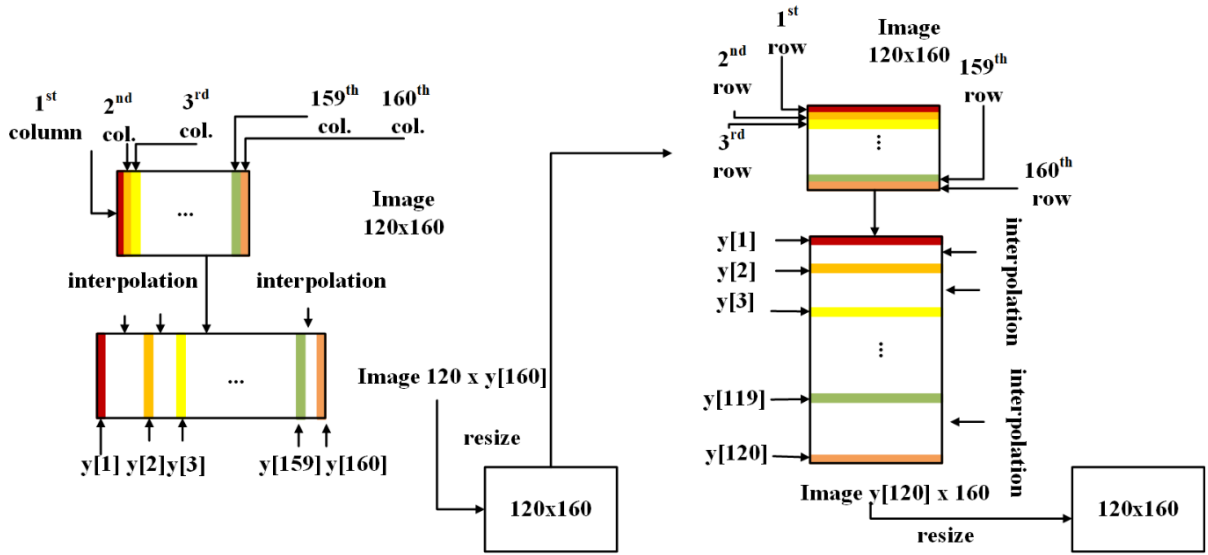

*Figure 13: Workflow of spatial stretching/contracting, illustrating the mapping of columns/rows to a new position based on y[j].*

Where $s$ is the length of the columns or rows of the original image depending on where the distortion is applied, $t[i]$ is a vector which includes all the integer values $[0, s-1]$ in ascending order and $\mathcal{U}(a, b)$ is the Uniform distribution between $a$ and $b$.

---
[7] https://github.com/C3Imaging/Deep-Learning-Techniques/tree/Off_axis_Iris

The first column/row of the original image is mapped on the first column/row of the stretched image. The following columns/rows are then mapped in the position determined by the value of $y[j]$. Depending on which is the desired direction for stretching the image, equation (5) or (6) is used in calculating $y[j]$. The combination of (5) and (6) makes it possible to stretch the images in four main directions. If (5) is used for mapping the columns and the rows, the image will be stretched in the right and down direction. In case (6) is used for both columns and rows, the image is stretched at left and up. Using (5) when mapping the columns and the (6) when mapping the rows of the image, will result in stretching the image to the right and up direction. Finally mapping the columns using (6) and the rows using (5), the image will be stretched to the left and down direction. Each direction has the same probability of being selected when the image is stretched. For each distortion and each direction, the amount of stretching applied to the images differs on every occasion as well as the volume of the stretching applied to the columns and the rows of the image is different, so that variation is injected to the augmented dataset. The stretching is applied at first to the columns of the image. The void spaces that are created, are interpolated with a weighted nearest neighbour method, which is explained by the following equations:

$$c[i] = \frac{\frac{f(y[j])}{i-y[j]} + \frac{f(y[j+1])}{y[j+1]-i}}{\frac{1}{i-y[j]} + \frac{1}{y[j+1]-i}}, j \in [1,160], i \in (y[j], y[j+1]) \qquad (9)$$

Where $f(x)$ is a function that returns the values of the $x^{th}$ column/row, $c[i]$ represents the values of the $i^{th}$ column/row of the stretched image. The values of $y[j]$ and $y[j+1]$ are the positions where the columns or rows of the stretched image have values and the columns/rows that need to be interpolated are located between these two positions. Finally, the image is contracted as the image is resized to the original resolution $[120 \times 160]$ using bicubic interpolation. The same process is then applied to the rows of the image. The same workflow is used for the ground truth segmentation map in order to obtain the segmentation map for the augmented sample. The described workflow for the spatial stretching/contracting of an image is illustrated in Figure 14. Applying spatial stretching/contracting results in an iris region that is not located in the centre of the image and with non-circular iris-pupil structures, as shown in Figure 15 which is a usual case in iris images acquired from a user-facing camera on AR/VR headsets.

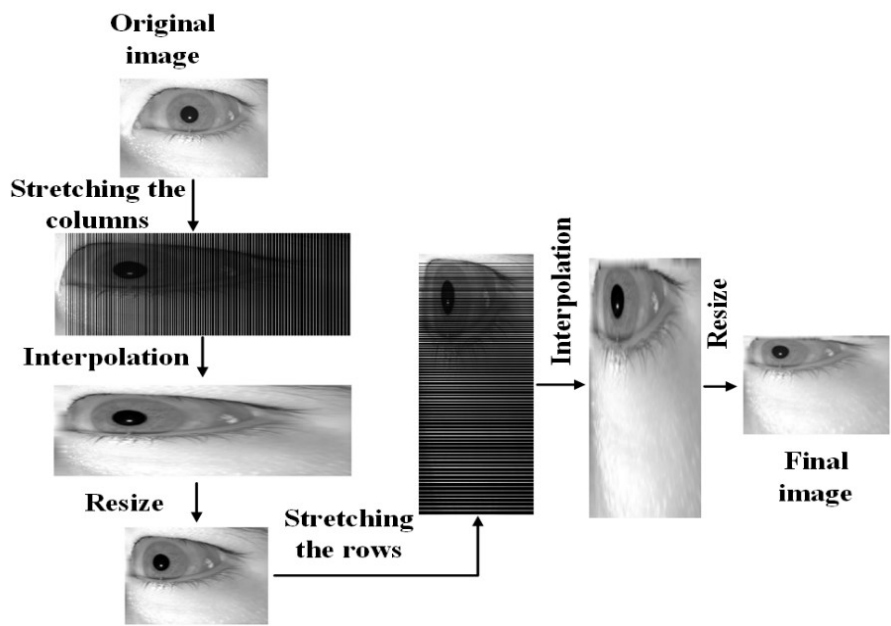

*Figure 14: Workflow of spatial stretching /contracting. For this transformation the equations (6) was used to map the columns and the rows of image and direct the image in up and left direction.*

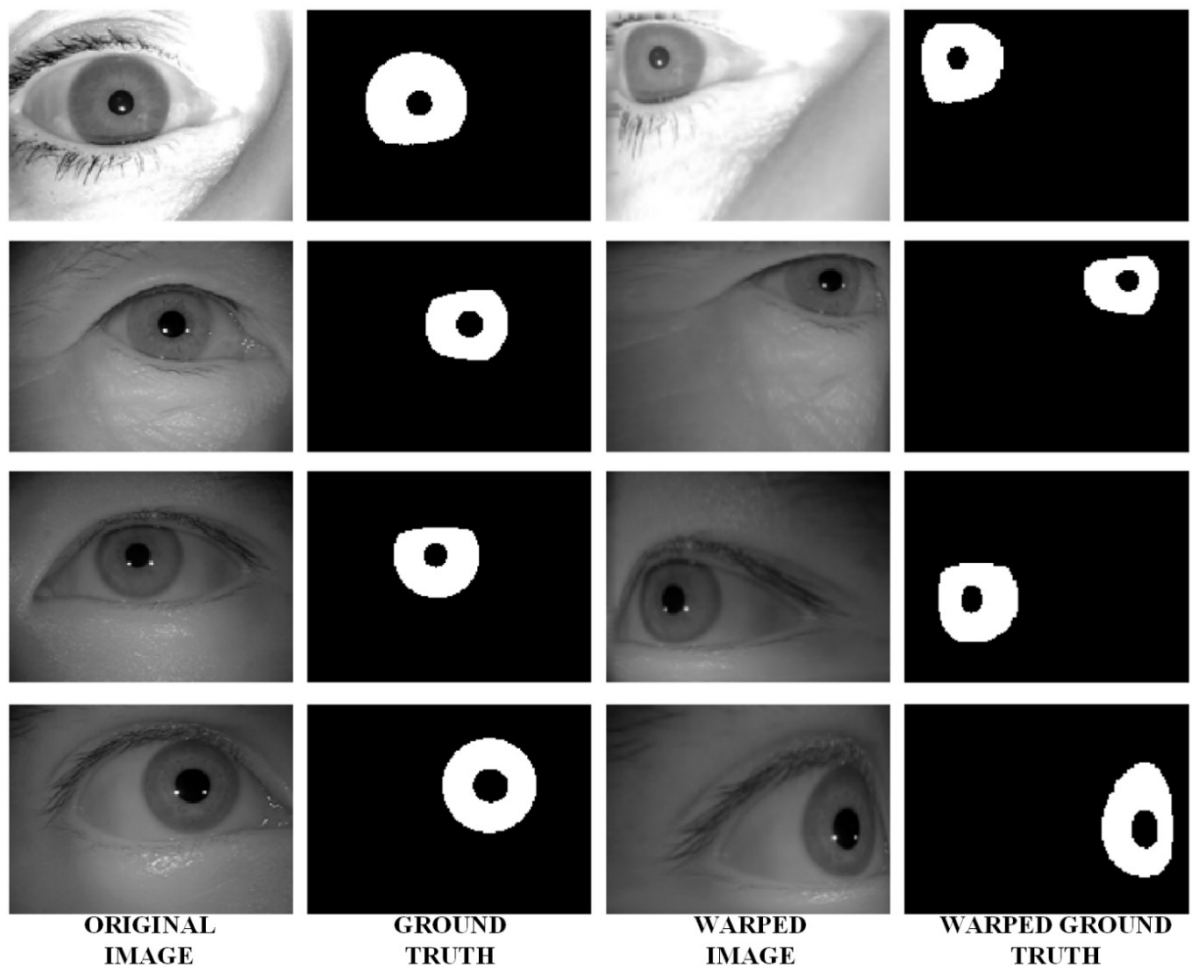

*Figure 15: Spatial stretched/contracted (warped) samples and their corresponding segmentation map.*

### 2.3.2.2. Augmentation 5: Image Tilting

A possible location of the camera used for capturing the iris images, as mentioned in the introduction and illustrated in Figures 1-2, will be below the eye. Therefore, the iris images should be representing samples which when captured, the camera is positioned below the eye level. To achieve that effect and also give an elliptical shape to the iris, in this second augmentation technique the samples are tilted in two directions: up and left, up and right.

A projective transformation is applied to the images. This transformation maps the top vertices of the image to a new pair of points as illustrated in Figure 16. The values from Figure 16, $a$, $b$, $c$, and $d$ are randomly generated between a range of values, so the image is tilted in the desired direction with variation. When the image is tilted up and left the values of $a$, $b$ are in $\mathcal{U}(0.15, 0.45)$, $c$ in $\mathcal{U}(0.9, 1)$ and $d$ in $\mathcal{U}(0, 0.1)$. When the image is tilted up and right, the values of $a$, $b$ are in $\mathcal{U}(0, 0.1)$, $c$ is in $\mathcal{U}(0.55, 1)$ and $d$ is in $\mathcal{U}(0.15, 0.45)$, where $\mathcal{U}(a, b)$ represents the Uniform distribution between $a$ and $b$. During this transformation as the image shrinks the interpolation used is the nearest-neighbour. The probability of the images being tilted in a direction between the two options (up and left / up and right) is the same.

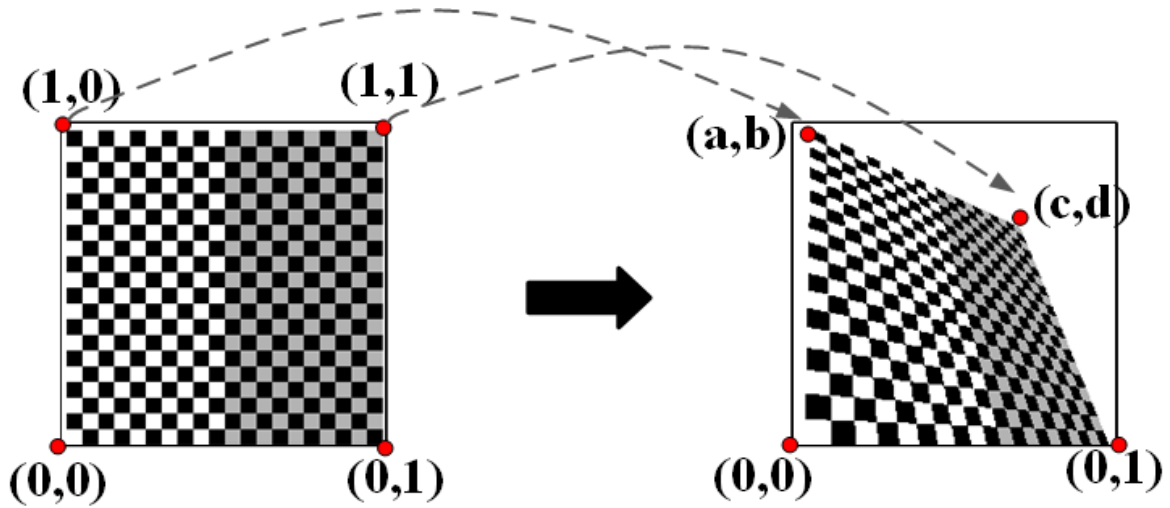

*Figure 16:Tilt transformation*

As shown in Figure 16, when the transformation is applied, while mapping the top vertices, to the *a*, *b*, *c* and *d* points, the image is compressed at the boundaries. Since the resolution of the image has to stay unchanged, the void spaces around boundaries should be filled to avoid sharp edges in the image. Since the void spaces are at the boundaries of the image, there isn't a direct way to apply interpolation. Therefore, the value from the edges of the tilted image is extended for each column up to the image boundary. The same process is applied to the image rows. After this process has been applied in both columns and rows, the average value is assigned to the void spaces. Finally, in order to smooth the interpolated areas of the image, a gaussian $3 \times 3$ filter with standard deviation $\sigma$ equal to 2 is applied to this region.

The described workflow for tilting an image is shown in Figure 17, and in Figure 18 samples are shown where the tilting transformation is applied. To obtain the segmentation map for the augmented samples, the same workflow is applied to the segmentation ground truth with the only difference being that the void spaces created are filled with black.

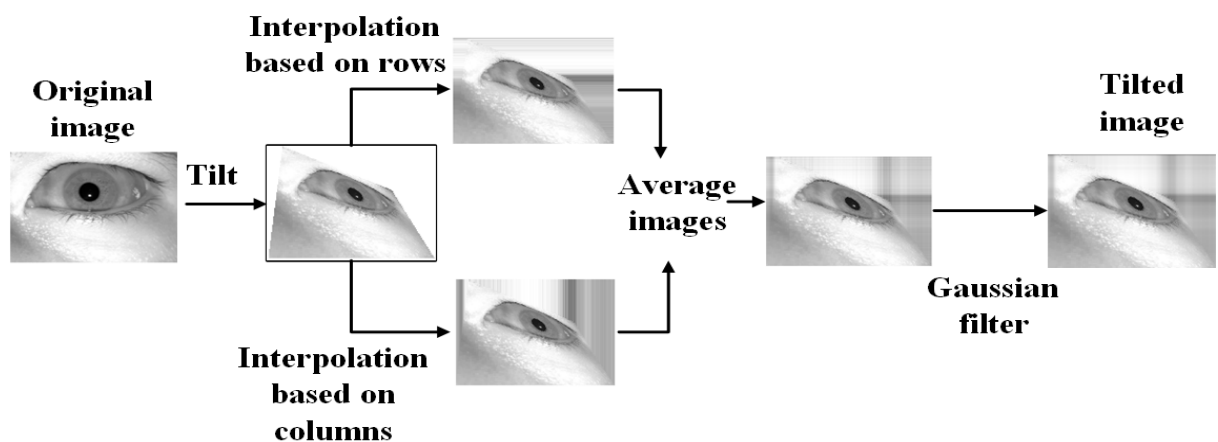

*Figure 17:Workflow of image tilting.*

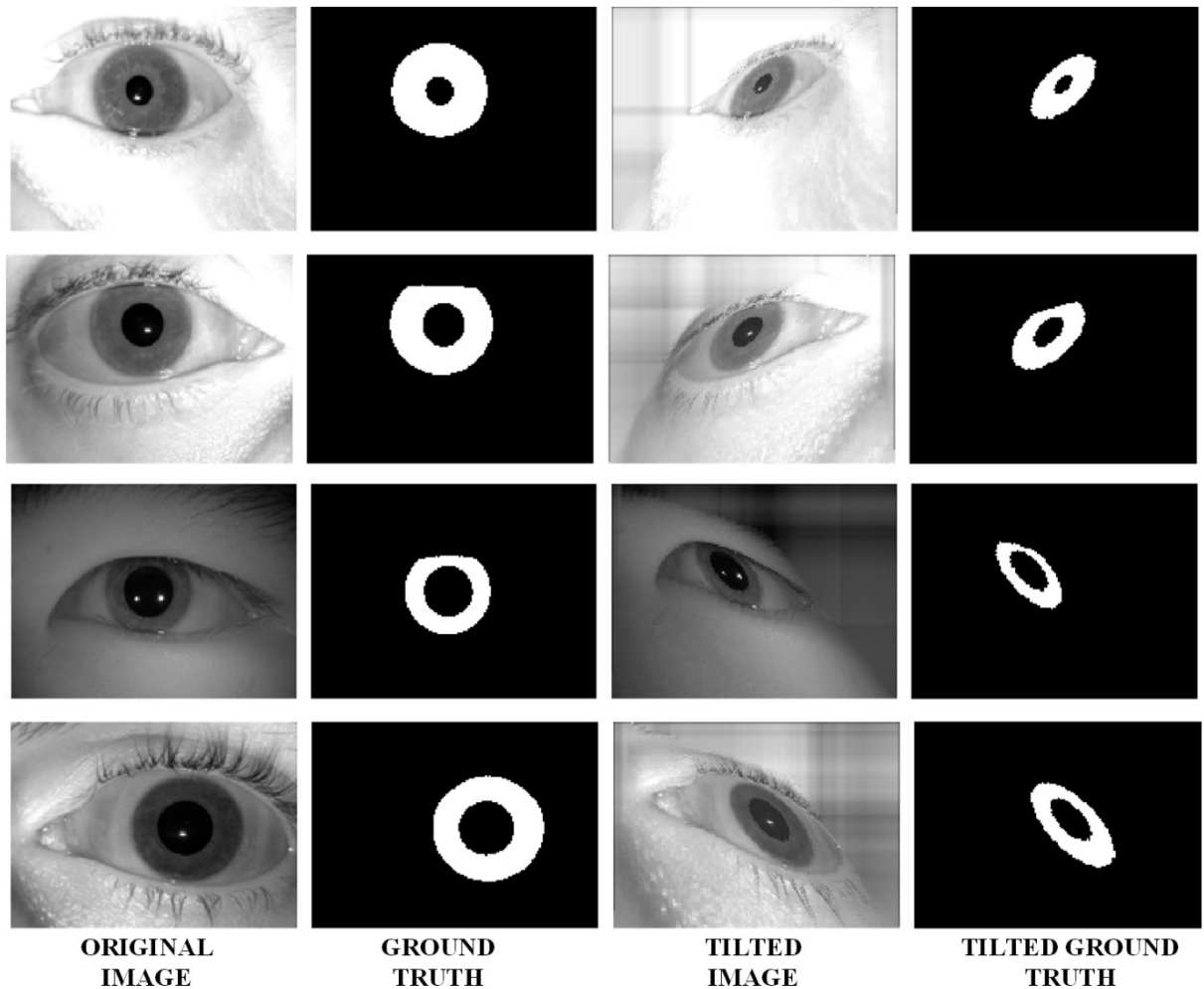

| ORIGINAL IMAGE | GROUND TRUTH | TILTED IMAGE | TILTED GROUND TRUTH |

*Figure 18:Tilted samples and their corresponding segmentation map.*

### 2.3.3. Dataset Preparation

#### 2.3.3.1. Workflow of Combining the Augmentations Techniques

The augmentation techniques are combined in various ways so that the dataset represents a generalized and realistic scenario and as a result the trained model can be robust and perform well in all the different conditions that one can encounter with iris images acquired from a user-facing camera on AR/VR device.

The augmentation techniques are mixed in three ways. Samples are augmented by only using the methods for simulating the off-axis, near-perspective iris images. At first, the spatial stretching/contracting transformation is applied to an image with 50% probability. In the next step, the tilting transformation is applied to the rest of the samples that the first transformation was not applied to. In addition to these, for an image that spatial stretching/contracting is applied to in the first step, there is a 50% probability that tilting is applied afterwards. In the second way, the samples are augmented using only the methods for simulating unconstrained conditions. At first, the contrast of all the iris images is modified as explained. Afterwards, all the images are passed through the motion filter, and finally, the technique used to introduce shadows is applied to the image. The probability that shadows are added to an image is 50%. Thirdly, the techniques from the two augmentation categories are combined. Initially, the techniques simulating the off-axis, near perspective iris images are applied to an image based on the augmentation workflow described above. Later the same image is processed using the techniques simulating unconstrained conditions, including contrast reduction, motion blurring, and shadowing. In Figure 19, the workflow explained is illustrated.

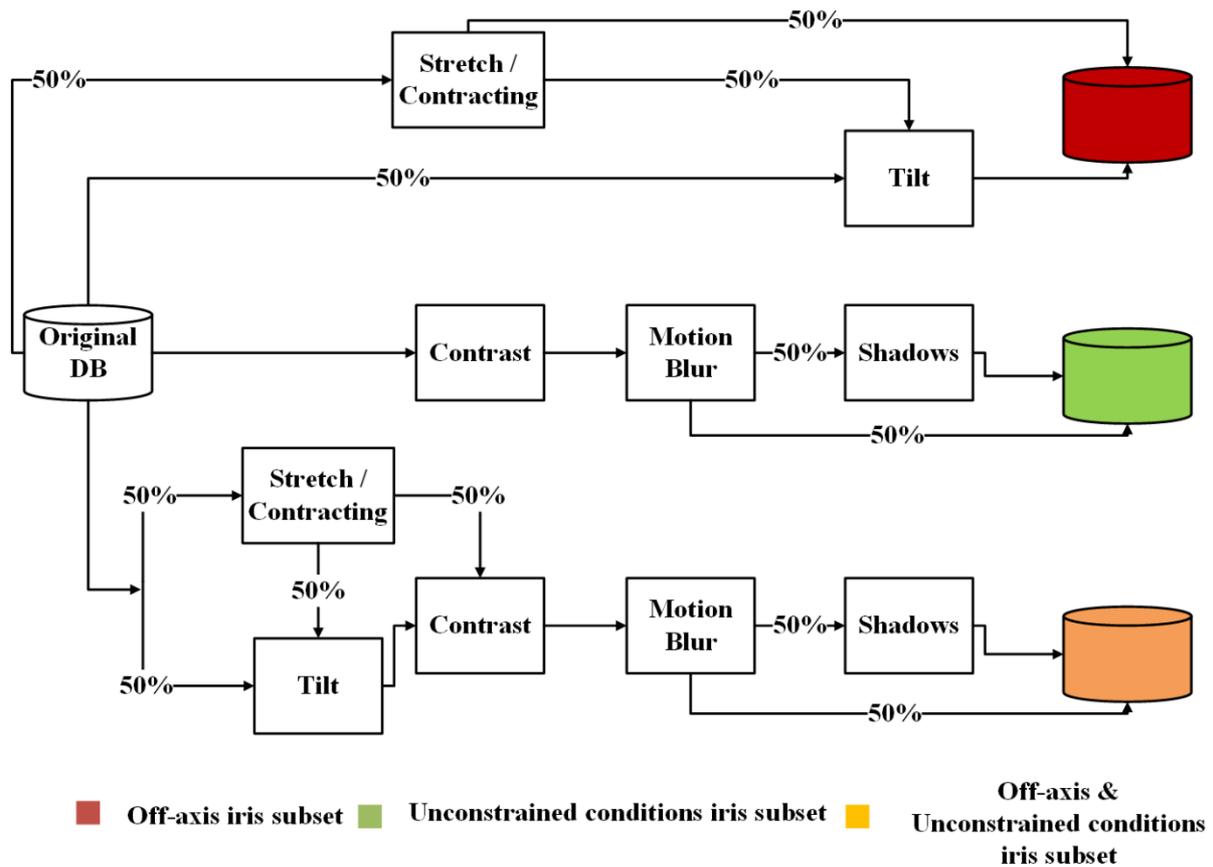

Figure 19:Workflow of augmentation techniques

The augmented samples simulate iris images captured using a user-facing camera on AR/VR device, frontal iris images affected by unconstrained conditions and AR/VR images affected by unconstrained conditions. Bath800 and CASIA Thousand are augmented with all three combinations of the augmentation techniques described. UBIRIS v2 was augmented only by using the augmentation techniques simulating iris images as represented by an AR/VR device. The samples of this dataset as mentioned previously are captured in unconstrained conditions, and therefore it will be redundant to make use of the augmentation techniques that simulate real-world conditions as they already exist in the dataset of UBIRIS v2.

### 2.3.3.2. Dataset Analysis

In this section a further analysis of the workflow used to combine the augmentation techniques is presented, in order to provide a better insight of the dataset created and used in this work.

As mentioned above, the workflow was designed in that way so that the dataset created to train the network represents a generalized and realistic problem. By using the three combinations of the augmentation techniques, three different subsets are created as shown in Figure 19, that form the main dataset used in this work. The first combination as described earlier uses only the augmentation techniques designed to simulate off-axis iris images. This process is used twice for each dataset creating the off-axis iris subset. The second combination uses only the augmentation techniques that simulate unconstrained conditions. This process is used once for each dataset consisting thus the unconstrained condition subset. Finally, the third combination, uses the augmentation techniques simulating the off-axis iris images and unconstrained conditions. This process is used twice for each dataset formulating the off-axis & unconstrained condition iris subset. Bath800 and CASIA Thousand combined consist of around 50.000 samples. With the use of the described workflow, 250.000 augmented samples are created. The off-axis iris subset is 100.000 samples, the unconstrained condition subset is 50.000 samples and 100.000 more samples from the unconstrained condition and off-axis iris subset. With the addition of the 50.000 original samples from Bath800 and CASIA Thousand, the final dataset used consist of 300.000 samples. In Table 1, a further analysis is presented describing the percentage of samples, with each augmentation technique or their combination, to the dataset.

*Table 1: Percentage (%) of images with each augmentation technique or combination in the dataset. In this table the augmentation techniques are referred as Contrast reduction: Contrast, Motion blur: Blur, Shadows: Shadows, Spatial stretching/contracting: Warp and Tilting: Tilt.*

| Augmentation Techniques | % of images | Dataset |
| --- | --- | --- |
| Contrast & Blur | ~8.5% | Unconstrained condition subset |
| Contrast & Blur & Shadows | ~8.5% | |
| Warp | ~8.5% | Off-axis subset |
| Tilt | ~16.5% | |
| Warp & Tilt | ~8.5% | |
| Warp & Contrast & Blur | ~4% | Off-axis & Unconstrained conditions subset |
| Tilt & Contrast & Blur | ~8.5% | |
| Tilt & Contrast & Blur & Shadows | ~8.5% | |
| Warp & Contrast & Blur & Shadows | ~4% | |
| Warp & Tilt & Contrast & Blur | ~4% | |
| Warp & Tilt & Contrast & Blur & Shadows | ~4% | |
| No augmentation | ~16.5% | Original subset |

Regarding the UBIRIS v2 as stated above, it consists of samples acquired in unconstrained conditions and therefore there is not a necessity of augmenting the samples with the augmentation techniques simulating unconstrained conditions. The samples of UBIRIS v2 are augmented only with the use of the augmentation techniques simulating off-axis iris images. This procedure is operated twice, creating the off-axis iris subset of UBIRIS v2 which along with the original samples are used in this work.

Finally, the element of randomness introduced in the augmentation techniques as well as at the way they are combined as explained in the workflow plays an important role to the augmentation process. One instance of that is that the direction of the shadowing, stretching or tilting is chosen randomly for each image. Also, the volume that an augmentation technique is applied to an image is chosen randomly between a range of values. Additionally, as illustrated in the workflow a sample is augmented with one or more augmentation techniques combined in different ways. These three approaches make it possible that a variety of conditions are introduced into the dataset leading to a generalized solution and each time producing unique samples with different characteristics and distributions. Examples of augmented samples with the use of all three different workflow combinations and their corresponding ground truth are in given in Figure 20.

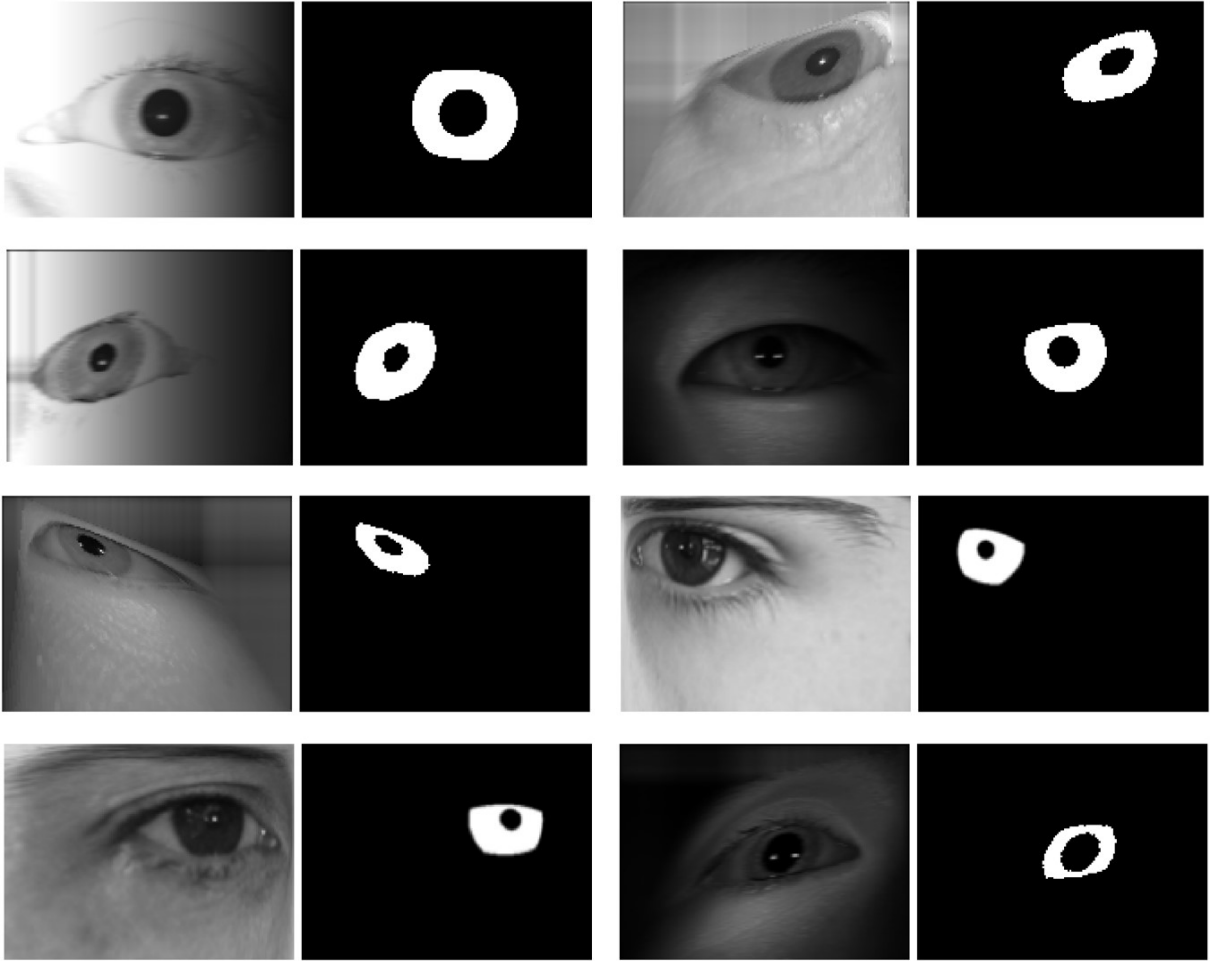

*Figure 20: Augmented samples and their corresponding ground truth.*

# 3. Network Design & Training

In this section the design of the network is presented along with a detailed comparison of its complexity with other CNN methods designed for the iris segmentation task followed by the procedure of training and fine-tuning.

## 3.1. Network Design

For the segmentation task, a fully convolutional network inspired by [43] is used, consisting of 10 layers. The network starts with a $3 \times 3$ kernel mapping the input (1 channel) on the first convolutional hidden layer which consists of 32 channels using a rectified linear unit (ReLu) as an activation function. The kernel size remains the same throughout the hidden convolutional layers, as well as, the number of channels and their activation function. Finally, at the output layer (1 channel), the kernel size is $3 \times 3$, but in this layer, the sigmoid activation function is used. Pooling layers were not used as it was observed that the performance of the network's output was decreasing. The design of the network is illustrated in Figure 21.

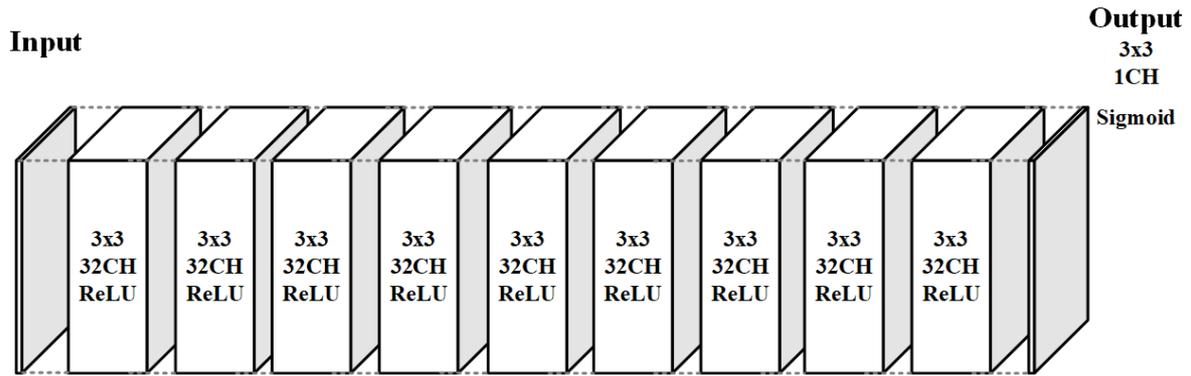

*Figure 21: Network Design*

## 3.2. Complexity Comparison of CNNs for Iris Segmentation

In this section, the complexity of several CNNs for iris segmentation will be compared with the proposed method. When referred to the complexity of a CNN, the main characteristics that one shall investigate is the number of parameters, the memory requirements for storing the parameters and the number of multiply-accumulate operations (MAC).

It is common practise that when an architecture of CNN is designed, that deep and large structures are favoured thus increasing the possibility of solving the investigated problem or promise higher performance from a smaller size CNN. Selecting a CNN with a deeper structure rather than a more compact structure, comes with some drawbacks such as increased training and execution time as well as generous memory requirements. There are cases, such as the proposed CNN, where a low complexity network can produce similar results as a high complexity network and as extension make it feasible to eliminate the downsides of a large CNN.

The proposed CNN consist of less than 75k parameters, requiring only 0.28MB of memory to store the parameters and 1426.64M MAC for an input image with dimensions $[120 \times 160 \times 1]$ $[width \times height \times channels]$. The SPDNN [43] consist of more than 1M parameters, requiring 35.26MB to store them and 13536.22M MAC for a smaller input image of dimensions $[98 \times 128 \times 1]$. Another high performance deep learning method, MFCNs [72] is of high complexity and memory requirements, with 21M parameters, needing 82.56MB of memory to store them. The input dimension and MAC in this structure are not specified as the input image dimension is not fixed and the number of MAC is related to the dimension of the input image. The complexity characteristics of the methods mentioned are shown in Table 2.

In this section is presented the low complexity proposed network, with reduced memory requirements resulting into a more efficient solution which is compatible for deployment in embedded applications such as AR/VR headsets. Furthermore, in the evaluation of the proposed network in section 5, is demonstrated that the low complexity network proposed in this work can obtain high performance iris segmentation results in both off-axis and frontal samples. The proposed CNN is outperforming other methods in segmenting off-axis iris images. Also, despite the fact that the network is designed for segmenting off-axis iris images, the results reported in segmenting frontal iris images are comparable to the-state-of-the-art SPDNN method of higher complexity and memory requirements.

*Table 2:Complexity of CNNs for Iris Segmentation*

| Metrics | Methods | | |
|---|---|---|---|
| | *Proposed Method* | *SPDNN* | *MFCNs* |
| Total no. parameters | 74.593 | 1.101.851 | 21.643.596 |
| Parameters size | 0.28MB | 35.26MB | 82.56 MB |
| Input size dimensions (width × height × channels) | 120 × 160 × 1 | 96 × 128 × 1 | N/A |
| Total MAC | 1426.64M | 13536.22M | N/A |

## 3.3. Training and Fine-tuning

### 3.3.1. Training

The network is trained on the original and augmented samples of Bath800 and CASIA Thousand. The dataset is divided 70% for the training set, 20% for validation set and 10% for the test set.

The training was carried out in TensorFlow library. The Mean Squared Error is used as the loss function. The Gradient Descent with Adaptive Moment Estimation (Adam) is used, with a learning rate of 1e-4, beta1 and beta2 equal to 0.9 and 0.999 respectively, to optimize the loss function. The training is done on a desktop computer with Nvidia GTX 1080 GPU.

### 3.3.2. Fine-tuning

In this section the process of fine-tuning the original network with the UBIRIS v2 dataset is described. Fine-tuning is a concept of transfer learning. Transfer learning is a machine learning technique, where knowledge gain during training in one type of problem is used to train in another related task or domain.

The proposed model was trained on the augmented and original samples from Bath800 and CASIA Thousand. UBIRIS v2 differs from the other datasets in the fact that it consists of visible iris image while Bath800 and CASIA Thousand are taken in NIR domain. Obtaining high-performance segmentation results in visible iris samples requires training a new model from the beginning or either fine-tune a pre-trained network on a dataset with visible samples. As UBIRIS v2 is a small dataset, training a new model is not possible, therefore fine-tuning the parameters of the pre-trained network is more functional. The network is trained on NIR iris samples and therefore it is excepted that the network transfers the information and tune the parameters on the UBIRIS v2 samples, as the context of the task and the datasets are similar.

Regarding the specifics of fine-tuning, the network is fine-tuned on the augmented and original samples of UBIRIS v2. The dataset is divided 70% for the training set, 20% for validation set and 10% for the test set. The training was carried out in TensorFlow library. The Mean Squared Error is used as the loss function. The Gradient Descent with Adaptive Moment Estimation (Adam) is used, with a learning rate of 5e-5, beta1 and beta2 equal to 0.9 and 0.999 respectively, to optimize the loss function. The tuning is done on a desktop computer with Nvidia GTX 1080 GPU.

# 4. Results

The input of the network is a grayscale iris image of 1 channel with dimensions [120 × 160]. The output of the network is a grayscale segmentation map with values between [0,1] and of the same size and channels as the input. The binary segmentation map is obtained by using a thresholding technique, where the values bigger than the threshold are shifted to 1 and the others to 0. The threshold value 0.55 is used in this work for the Bath800, CASIA Thousand which are datasets containing NIR images. Regarding UBIRIS v2 which contains visible samples, after fine-tuning the network to the dataset, the threshold with value 0.4 is selected. The output of the proposed model for the different datasets are shown in Figures 22-24.

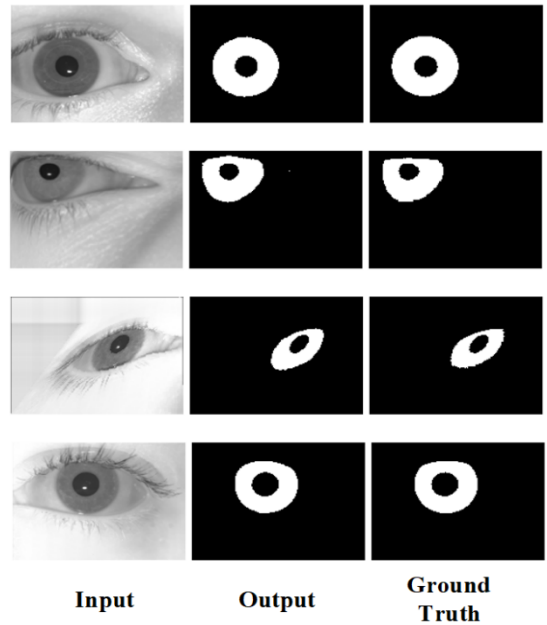

Figure 22: Output of the network for the augmented off-axis and original samples of Bath800.

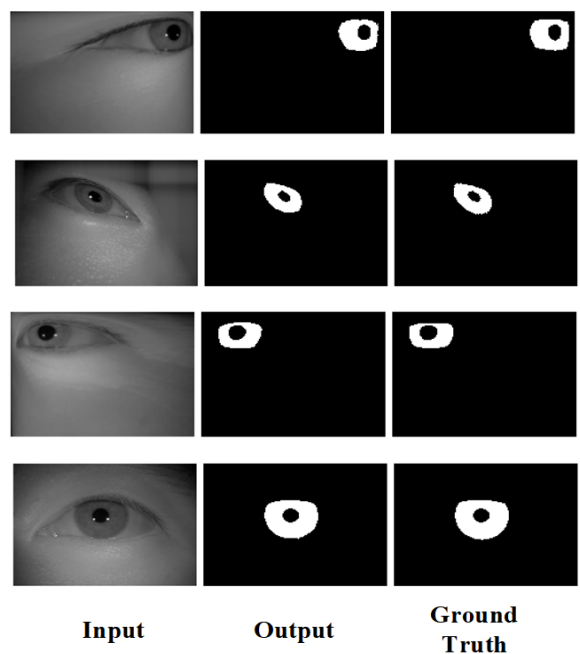

Figure 23: Output of the network for the augmented off-axis and original samples of CASIA Thousand.

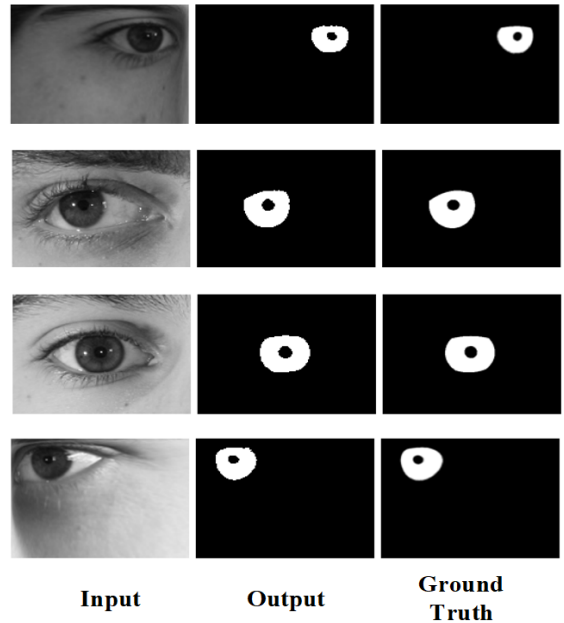

*Figure 24:Output of the network for the augmented off-axis and original samples of UBIRIS v2.*

# 5. Evaluation

Several metrics are used to evaluate the proposed method and conduct a detailed comparison with several segmentation methods of the literature. The metrics used in this work are: accuracy, sensitivity, specificity, precision, NPV and F1-score. More information about these metrics can be found in [43]. Two main experiments have been used to evaluate the performance of the proposed network:

1) Evaluate the proposed network on the off-axis augmented samples:

The network is trained on the original and augmented samples of Bath800 and CASIA Thousand. The network is tested on the off-axis augmented samples of Bath800, CASIA Thousand and UBIRIS v2. These are the off-axis subset and off-axis with unconstrained condition subset for Bath800 and CASIA Thousand and the off-axis subset for UBIRIS v2, as described in the section 2.3.3. In continuance it is compared with the segmentation results on these samples from the methods: SPDNN [43], IrisSeg [84] and OSIRIS [85]. The test set of the augmented samples is used to test the network and the other methods.

2) Evaluate the network on the original samples from the datasets:

The network is tested on the original samples of Bath800, CASIA Thousand and UBIRIS v2, which consist of frontal iris samples. The test set of these datasets are used for testing the proposed method. The results of the proposed network are compared extensively with the state-of-the-art SPDNN on the Bath800, CASIA Thousand and UBIRIS v2. Furthermore, the results of the network are compared with the available results from other segmentation methods of the literature.

The results presented on UBIRIS v2 are the results of the original network after tuning.

## 5.1. Evaluation and Comparison on off-axis Augmented Samples

In this section the proposed method is tested on the off-axis augmented samples. These samples are the combination of the off-axis subset and the off-axis with unconstrained condition subset for Bath800 and CASIA Thousand and the off-axis iris subset for UBIRIS v2. The test sets from the datasets are used for the testing stage.

### 5.1.1. Evaluation

The proposed network produces high performance results in the datasets Bath800 and CASIA Thousand. This is expected as the network is trained on them. On UBIRIS v2 the network is able to provide accurate segmentation results but is not able to perform at the same level as on Bath800 and CASIA Thousand. The samples of UBIRIS v2 are taken in visible spectrum and therefore the distribution differs. The proposed network with tuning is able to produce high segmentation results showing that the CNN is able to adopt to a similar task but with different distribution.

In the datasets that the network is trained on, the accuracy and the F1-score and sensitivity measurements are higher showing high quality in returning true results and more consistent segmentations in comparison with UBIRIS v2. The same applies with the sensitivity and NPV metrics showing that the network is able to rule-out non-iris pixels more effectively in the trained datasets than the dataset that it was tuned on. Although, the network has higher performance in precision and specificity on the UBIRIS v2 dataset, showing greater capability in returning real iris pixels in the UBIRIS v2 dataset rather than the Bath800 and CASIA Thousand. The results are shown in Tables 3-5.

### 5.1.2. Comparison with SPDNN, IrisSeg and OSIRIS

The proposed method is designed for segmenting low quality off-axis iris images as acquired from an AR/VR device. The proposed method is compared with the SPDNN, IrisSeg and OSIRIS on the test set of the augmented off-axis samples. The selection of these algorithms is based on their availability. Furthermore, the SPDNN is a state-of-the-art segmentation method specialized on low quality iris images and IrisSeg and OSIRIS are well-established methods with high performance in the iris segmentation task. The SPDNN is trained on the original and augmented samples of Bath800 and CASIA Thousand and tuned on UBIRIS v2. The augmented samples used in their work are representing unconstrained scenarios. The SPDNN is a network with high capacity and large number of parameters as analysed earlier.

The SPDNN when tested on the off-axis augmented samples is able to provide overall good results in accuracy and specificity and average results in precision. The performance of the SPDNN is low in the sensitivity and F1-score measurements. The proposed network is outperforming the SPDNN in all the evaluation metrics showing higher results and ability to segment off-axis iris samples as appear when acquired from a user-facing camera on AR/VR device. In regard to IrisSeg and OSIRIS there not able to provide high segmentation results for the augmented off-axis samples. The low performance results of IrisSeg and OSIRIS are due to the fact that the augmented samples that used are challenging as they simulate off-axis iris images in unconstrained conditions. In addition, IrisSeg and OSIRIS were not able to provide a segmentation in many cases. The results included for IrisSeg and OSIRIS are only for the images that the algorithms were able to provide a segmentation. The results are given in Tables 3-5.

Table 3:Comparison of the proposed method with other segmentation methods on the off-axis augmented samples of Bath800. A higher value for $\mu$ and lower for $\sigma$ is desired.

| Metrics | | Bath800 | | | |
| --- | --- | --- | --- | --- | --- |
| | | *Proposed Method* | *SPDNN* | *IrisSeg* | *OSIRIS* |
| Accuracy | $\mu$ | 99.22% | 97.03% | 96.10% | 95.86% |
| | $\sigma$ | 0.62% | 1.96% | 3.53% | 2.80% |
| Sensitivity | $\mu$ | 92.98% | 58.71% | 67.26% | 62.16% |
| | $\sigma$ | 8.7% | 38.04% | 21.82% | 35.72% |
| Specificity | $\mu$ | 99.62% | 99.15% | 98.02% | 98.00% |
| | $\sigma$ | 0.38% | 0.86% | 3.53% | 2.37% |
| Precision | $\mu$ | 93.97 | 80.34% | 75.88% | 67.68% |
| | $\sigma$ | 7.41% | 19.32% | 21.18% | 24.11% |
| NPV | $\mu$ | 99.52% | 97.74% | 97.79% | 97.60% |
| | $\sigma$ | 0.57% | 2.14% | 1.72% | 2.24% |
| F1-Score | $\mu$ | 93.21% | 59.90% | 68.63% | 59.54% |
| | $\sigma$ | 7.70% | 35.76% | 19.51% | 31.78% |

Table 4:Comparison of the proposed method with other segmentation methods on the off-axis augmented samples of CASIA Thousand. A higher value for $\mu$ and lower for $\sigma$ is desired

| Metrics | | CASIA Thousand | | | |
| --- | --- | --- | --- | --- | --- |
| | | *Proposed Method* | *SPDNN* | *IrisSeg* | *OSIRIS* |
| Accuracy | $\mu$ | 99.40% | 97.75% | 96.7% | 95.81% |
| | $\sigma$ | 0.56% | 1.66% | 5.52% | 2.49% |
| Sensitivity | $\mu$ | 90.64% | 49.36% | 69.67% | 36.34% |
| | $\sigma$ | 11.14% | 43.15% | 27.13% | 38.40% |
| Specificity | $\mu$ | 99.77% | 99.43% | 97.90% | 98.60% |
| | $\sigma$ | 0.29% | 0.9% | 5.62% | 2.13% |
| Precision | $\mu$ | 94.17% | 75.89% | 74.37% | 48.42% |
| | $\sigma$ | 7.87% | 28.26% | 27.97% | 34.48% |
| NPV | $\mu$ | 99.59% | 98.27% | 98.63% | 97.07% |
| | $\sigma$ | 0.49% | 1.64% | 1.19% | 2.14% |
| F1-Score | $\mu$ | 91.93% | 49.40% | 69.00% | 35.83% |
| | $\sigma$ | 9.66% | 41.44% | 26.72% | 34.31% |

Table 5: Comparison of the proposed method with other segmentation methods on the off-axis augmented samples of UBIRIS v2. A higher value for $\mu$ and lower for $\sigma$ is desired

| Metrics | | UBIRIS v2 | | | |
|---|---|---|---|---|---|
| | | **Proposed Method** | *SPDNN* | *IrisSeg* | *OSIRIS* |
| Accuracy | $\mu$ | 98.83% | 97.94% | 87.17% | 92.96% |
| | $\sigma$ | 1.16% | 1.84% | 9.10% | 5.31% |
| Sensitivity | $\mu$ | 83.89% | 60.17% | 27.06% | 24.11% |
| | $\sigma$ | 10.48% | 34.20% | 23.51% | 29.04% |
| Specificity | $\mu$ | 99.77% | 99.75% | 91.03% | 97.58% |
| | $\sigma$ | 0.46% | 0.73% | 9.30% | 4.15% |
| Precision | $\mu$ | 95.26% | 93.78% | 24.31% | 39.11% |
| | $\sigma$ | 9.87% | 15.51% | 29.21% | 38.34% |
| NPV | $\mu$ | 98.94% | 98.01% | 94.97% | 95.15% |
| | $\sigma$ | 1.12% | 1.88%% | 4.22% | 4.33% |
| F1-Score | $\mu$ | 88.72% | 66.35% | 21.58% | 23.58% |
| | $\sigma$ | 10.62 | 35.49% | 22.50% | 29.66% |

## 5.2. Evaluation and Comparison on the frontal iris-region Samples

In this section the proposed method is evaluated and compared on the frontal original samples of Bath800, CASIA Thousand and UBIRIS v2, which consist of frontal iris samples. It is worthwhile to note that the proposed technique is designed for segmenting off-axis consumer level iris images. Despite that, the experiments below are carried out in order to conduct a fair comparison with the other methods on frontal images. Meanwhile the proposed method is giving the best results on segmenting the augmented off-axis samples.

### 5.2.1. Evaluation on the frontal iris-region samples

The proposed network is now tested on the original samples from Bath800, CASIA Thousand and UBIRIS v2. For this procedure the test sets of the datasets are used.

Similar outcomes with the one's on the evaluation of the proposed method on the off-axis iris samples are found in the evaluation of the original samples. The proposed network has higher performance in the datasets that the network is trained on, Bath800 and CASIA Thousand. Lower performance is reported on UBIRIS v2. The network accomplishes high accuracy results in all datasets showing that has high quality in returning true results. Moreover, in all datasets it returns high values in specificity and precision, meaning that the model performs well returning iris pixels. The sensitivity metric on Bath800 and CASIA Thousand is high, showing the ability of the model in ruling out non-iris pixels accurately while in UBIRIS v2 the same metric has average performance. The same applies to the F1-score measurement showing that the network produces more consistent segmentations, both in finding iris and non-iris pixels in the datasets Bath800 and CASIA Thousand compared to UBIRIS v2. In the Table 6 the results of the proposed network on the test sets of the original samples from Bath800, CASIA Thousand and UBIRIS v2 are presented.

*Table 6: Testing the proposed method on the original samples of several datasets.*

| Metrics | | Proposed Method | | |
|---|---|---|---|---|
| | | *Bath800* | *CASIA Thousand* | *UBIRIS v2* |
| Accuracy | $\mu$ | 99.13% | 99.50% | 98.92% |
| | $\sigma$ | 0.52% | 0.36% | 0.67% |
| Sensitivity | $\mu$ | 94.90% | 94.67% | 88.38% |
| | $\sigma$ | 6% | 4.33% | 9.29% |
| Specificity | $\mu$ | 99.56% | 99.86% | 99.71% |
| | $\sigma$ | 0.47% | 0.16% | 0.39% |
| Precision | $\mu$ | 95.67% | 97.39% | 96.33% |
| | $\sigma$ | 6.33% | 2.83% | 7.22% |
| NPV | $\mu$ | 99.49% | 99.63% | 99.10% |
| | $\sigma$ | 0.45% | 0.34% | 0.60% |
| F1-Score | $\mu$ | 95.17% | 95.94% | 91.46% |
| | $\sigma$ | 5.43% | 2.89% | 9.63% |

### 5.2.2. Comparison with the SPDNN

The SPDNN is a sophisticated network, with state-of-the-art results in the iris segmentation task. Now as mentioned earlier the SPDNN it was trained on samples of Bath800 and CASIA Thousand and tuned on the UBIRIS v2, as is the proposed method. The SPDNN is of high complexity with 14 times more number of parameters when compared to the proposed network. This is an aspect that should be considered in the comparison between these segmentation methods. Also, as mentioned earlier the proposed network is designed for segmenting off-axis iris images as captured by a user-facing camera on AR/VR device. The numerical results of the SPDNN [43] performance are reported as presented in their work .

#### 5.2.2.1. Comparing results on Bath800, CASIA Thousand and UBIRIS v2

The proposed method shows higher accuracy than the SPDNN in the Bath800 dataset which implies better quality in returning true results. The performance in specificity of the proposed method is also higher than the SPDNN. However in the precision metric the SPDNN is performing better. That shows that both can perform well in returning iris pixels, with not one method being better than the other. The same applies in the ability of the methods in ruling out non-iris pixels, as in NPV the proposed method is performing better than the SPDNN while the SPDNN shows higher results from the proposed method in the sensitivity metric. On the other hand, a small advantage of the SPDNN over the proposed method is in the F1-score showing a better efficiency. Overall in Bath800 dataset there isn't a clear advantage of one method over the other as the performance in the metrics is divided with the differences between them either in favour or against them being marginal. The proposed network is performing comparable with the SPDNN in the Bath800 dataset.

In regards with the CASIA Thousand dataset, the SPDNN shows a small advantage over the proposed method. The proposed method performs better in the specificity and precision metrics showing higher quality in returning iris pixels than the SPDNN. In the rest of the evaluation metrics the SPDNN is performing better than the proposed method. Nonetheless, generally the differences in performance are marginal.

On UBIRIS v2, the SPDNN performs better than the proposed method. The proposed method is performing better only in the specificity and precision metrics showing that is better on returning iris pixels than the SPDNN. In some metrics such as accuracy and NPV the difference is marginal showing that the proposed method is almost as good as the SPDNN in returning true results and in ruling out non-iris pixels. In the rest of the metrics there is a slight difference between the two methods, showing that the SPDNN is able to adopt better to the dataset that the methods are tuned utilising thus the larger number of parameters of the SPDNN.

Overall, the proposed network and SPDNN performs similarly in the Bath800 and CASIA Thousand datasets, which are the datasets that were trained on. Therefore, showing that when trained the proposed network is comparable to the SPDNN despite that the complexity of the proposed network is at least an order of magnitude less than the SPDNN as analyzed earlier. In the UBIRIS v2, where the proposed network and the SPDNN are tuned, the proposed method shows high results but the SPDNN still outperforms it, showing the ability to adopt better to a different dataset distribution, utilizing the higher complexity of its structure. The comparison between the two methods is shown in Table 7.

*Table 7: Comparison between the proposed method and the SPDNN on the original samples from several datasets. Green colour shows a better performance. Yellow shows a marginal difference in the performance and Orange a noteworthy difference in performance. A higher value for $\mu$ and lower for $\sigma$ is desired*

| Metrics | | Bath800 | | CASIA Thousand | | UBIRIS v2 | |
|---|---|---|---|---|---|---|---|
| | | *Proposed Method* | SPDNN | *Proposed Method* | SPDNN | *Proposed Method* | SPDNN |
| Accuracy | $\mu$ | 99.13% | 98.55% | 99.50% | 99.71% | 98.92% | 99.30% |
| | $\sigma$ | 0.52% | 1.43% | 0.36% | 0.33% | 0.67% | 0.54% |
| Sensitivity | $\mu$ | 94.90% | 96.03% | 94.67% | 97.96% | 88.38% | 93.98% |
| | $\sigma$ | 6% | 4.76% | 4.33% | 2.95% | 9.29% | 9.45% |
| Specificity | $\mu$ | 99.56% | 99.10% | 99.86% | 99.82% | 99.71% | 99.62% |
| | $\sigma$ | 0.47% | 1.07% | 0.16% | 0.20% | 0.39% | 0.48% |
| Precision | $\mu$ | 95.67% | 96.05% | 97.39% | 97.13% | 96.33% | 94.88% |
| | $\sigma$ | 6.33% | 4.46% | 2.83% | 3.10% | 7.22% | 5.40% |
| NPV | $\mu$ | 99.49% | 99.05% | 99.63% | 99.87% | 99.10% | 99.60% |
| | $\sigma$ | 0.45% | 1.49% | 0.34% | 0.28% | 0.60% | 0.30% |
| F1-Score | $\mu$ | 95.17% | 95.93% | 95.94% | 97.50% | 91.46% | 93.90% |
| | $\sigma$ | 5.43% | 3.88% | 2.89% | 2.51% | 9.63% | 9.70% |

### 5.2.3. Comparison with state-of-the-art methods

In the following section the proposed method is compared to the most advance and state-of-the-art segmentation methods in the literature. First, accuracy over the challenging UBIRIS v2 dataset is compared with several methods. In continuance, it is evaluated and compared over CASIA Thousand and UBIRIS v2 in three important segmentation metrics: sensitivity, precision and F1-score with known segmentation methods.

#### 5.2.3.1. Comparison of accuracy on UBIRIS v2

The accuracy of the proposed method is compared with state-of-the-art segmentation methods over UBIRIS v2. The state-of-the-art segmentation methods used in the comparison are the following: SPDNN [43], MFCN and HCNN from [72],[86], [87],[88], [89],[90] and [91]. A brief description of these methods can be found in [43]. The performance on accuracy of the UBIRIS dataset of the proposed method compared with the aforementioned state-of-the-art methods are presented in the Figure 25.

As illustrated the proposed method has the third best performance compared with the state-of-the-art segmentation methods. The two methods that are performing better are: the SPDNN of [43] and MFCN of [72]. Despite the fact that are performing better than the proposed method, one shall take into consideration the complexity of these CNNs. As analysed in the section 3.2, the SPDNN and the MFCN are of high complexity. The SPDNN consists of over a 1M of parameters requiring 35.26MB of memory to store them and the MFCN is estimated to consist of more than 21M parameters requiring 82.56 MB to store them. On the other hand, the proposed network is of low complexity with less than 75K parameters requiring only 0.28MB to store them.

Overall, the proposed method is the third best performing algorithm in the challenging dataset of UBIRIS v2 while is its complexity is at least an order of magnitude less than the two methods that outperforms it, making the proposed method more suited for deployment in embedded applications.

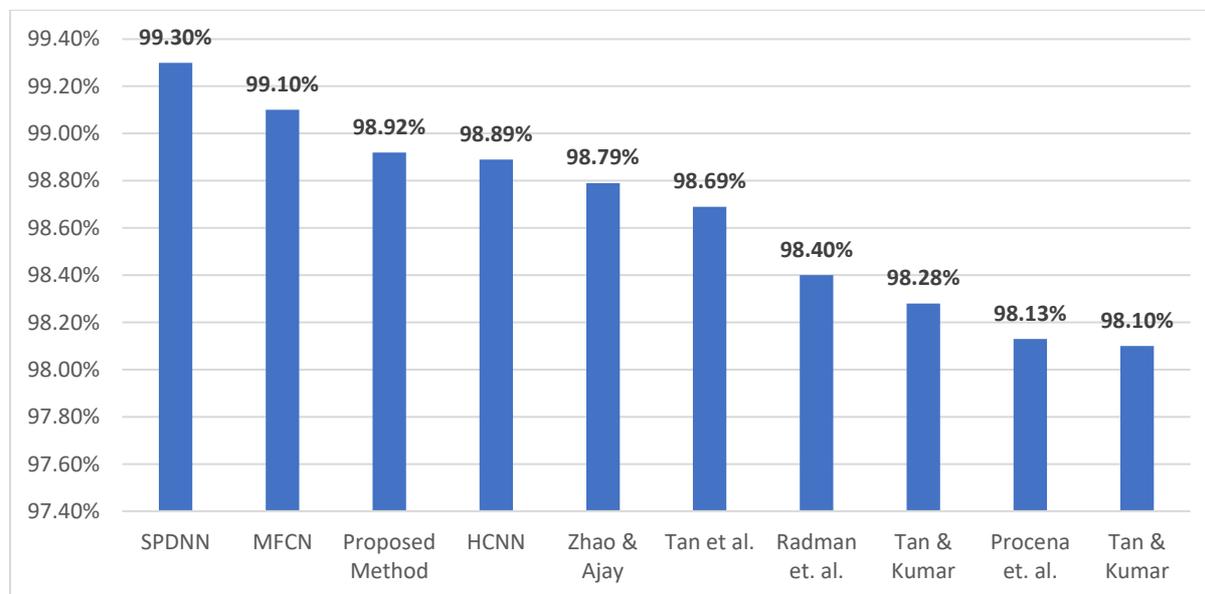

*Figure 25:Accuracy of the proposed method vs other methods over the original UBIRIS v2 dataset.*

#### 5.2.3.2. Comparison on CASIA Thousand and UBIRIS v2

In this section the proposed method is compared with other known segmentation methods over three important metrics: sensitivity, precision and F1-score. Sensitivity measures the ability of a model in ruling out non-iris pixels, while precision measures the ability of the model to detect true iris pixels. F1-score is the harmonic average of these two metrics. The segmentation methods include CAHT, GST, IFFP, OSIRIS, and WAHET. The comparisons are made over the CASIA Thousand and UBIRIS v2 original datasets. The numerical results are initially presented at [53]. The metrics for each presented algorithm are calculated comparing the algorithms results with the ground truth. The comparisons are illustrated in the Figures 26-27. On CASIA Thousand dataset, a high-quality dataset, is performing better than the other methods in precision and F1-score, while in the sensitivity metric only CAHT is giving better result than the proposed method. Moreover, on UBIRIS v2 where the samples are of low quality the proposed method gives higher results in all metrics compared to the other approaches. This shows that the proposed method although its designed for segmenting off-axis iris samples as represented by a user-facing camera on AR/VR device, is performing well on frontal iris samples of both high and low quality.

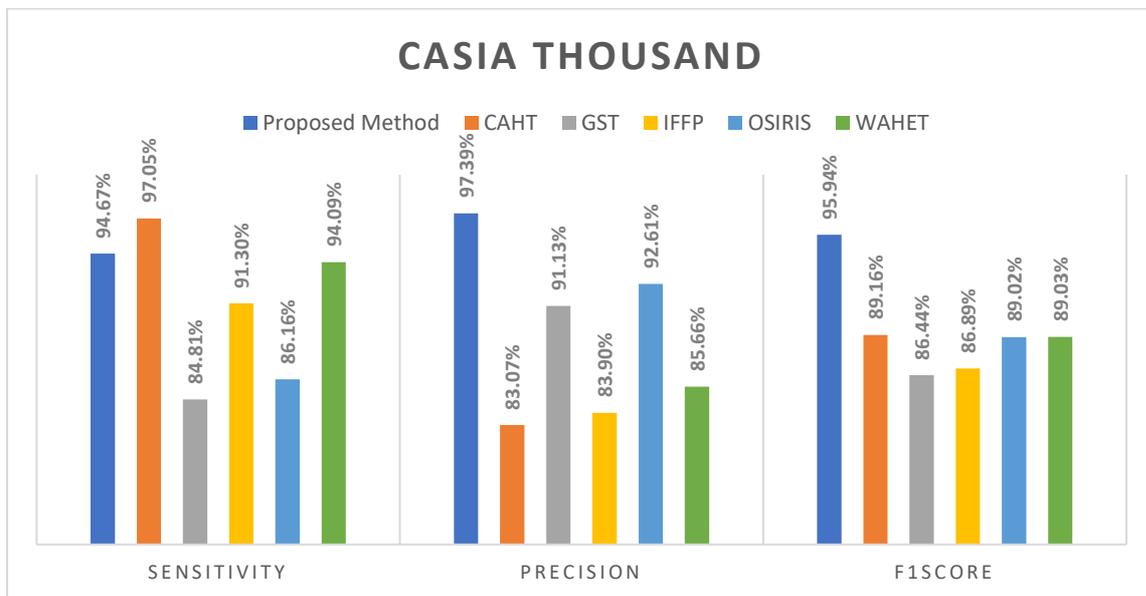

*Figure 26: Sensitivity, Precision, F1-score on the original samples of CASIA Thousand for the proposed method vs five other methods*

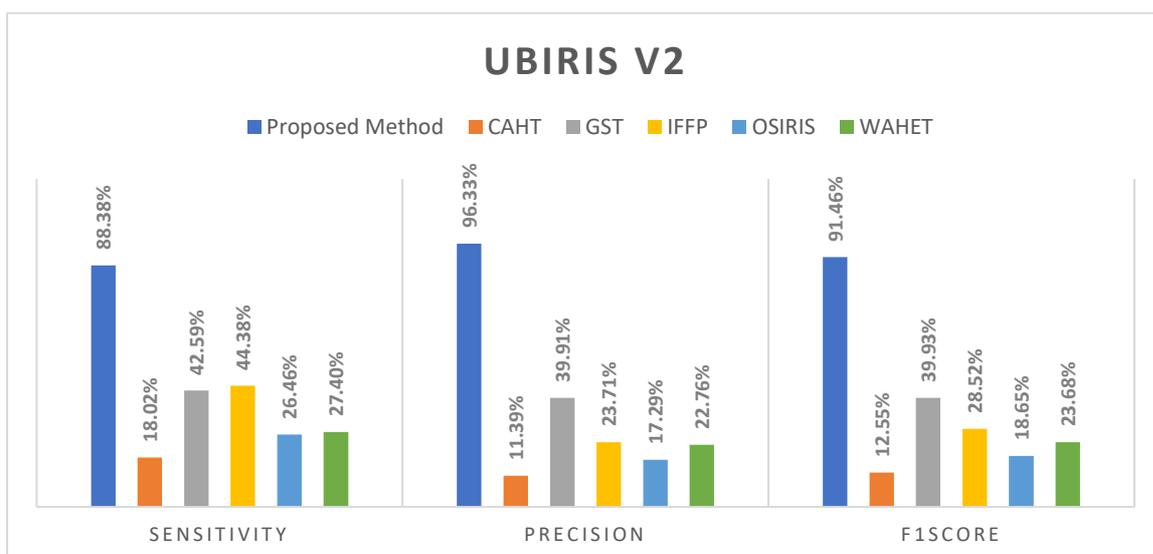

*Figure 27: Sensitivity, Precision, F1-score on the original samples of UBIRIS v2 for the proposed method vs five other methods*

# 6. Conclusion

In this paper the design for a low-complexity neural network targeted to segment off-axis iris regions as captured by user-facing cameras on wearable AR/VR headsets is presented. Advanced augmentation methods are provided to facilitate the training of this network from high-quality public datasets. The quality of segmentation achieved by this network is evaluated and compared with state-of-the-art methods both for off-axis and frontal iris regions.

The proposed network's complexity is at least an order of magnitude less than other CNNs specifically designed for the iris segmentation task. Also, it has the best performance on segmenting the augmented off-axis iris samples. Further, the segmentation performance of this network on frontal iris samples from several public datasets, is comparable with the SPDNN network proposed by [43] a state-of-the-art iris segmentation method. This performance is achieved even though the proposed network is of significantly smaller size and complexity and is trained for the task of segmenting off-axis iris samples. Due to its lightweight design and high performance in segmenting both off-axis and frontal iris samples and handling a range of input image qualities, the proposed network is well suited for general deployment on AR/VR devices.

Future work will focus on refinements in the network design and training/augmentation methodologies to improve performance on specific AR/VR headsets. As can be noted from the introduction, different devices will have user-facing cameras in a more limited set of locations and image acquisitions will be at varying NIR/wavelengths. In addition, the imaging pipeline on each camera module can have subtle effects on image quality.

Some practical examples of further research topics include developing an optimized CNN design based on SPDNN methods with a similar, or perhaps even smaller number of parameters that can achieve similar segmentation accuracy to our network. Another future research direction is to build some device-specific datasets with iris images captured by the user-facing camera on several state-of-the-art AR/VR headsets. This will enable evaluation of the proposed segmentation method on practical off-axis iris samples. (At present it is not possible to gain low-level access to the imaging systems on the available devices to capture continuous image streams, but we have opened some discussions with device manufacturers and such access will hopefully be available in the near future as these devices continue to enter mainstream adoption.)

It is also expected to extend this work to apply these improved segmentation techniques to a number of full iris recognition pipelines to evaluate its effects on the reliability and robustness of near-view, off-axis iris recognition. The main challenge here is that the only off-axis recognition pipeline that we are aware of is proprietary. Again, we expect other algorithms will appear in the near future and hopefully some of these will be open-source or provide at least API-level access to system developers.

**Acknowledgments**

This research is funded under the SFI Strategic Partnership Program by Science Foundation Ireland (SFI) and FotoNation Ltd. Project ID: 13/SPP/I2868 on Next Generation Imaging for Smart- phone and Embedded Platforms.

Portions of the research in this paper use the CASIA-IrisV4 collected by the Chinese Academy of Sciences' Institute of Automation (CASIA).